# Coding the Visual World: From Image to Simulation Using Vision Language Models

Sagi Eppel[1]

## Abstract

The ability to construct mental models of the world is a central aspect of understanding. Similarly, visual understanding can be viewed as the ability to construct a representative model of the system depicted in an image. This work explores the capacity of Vision Language Models (VLMs) to recognize and simulate the systems and mechanisms depicted in images using the Im2Sim methodology. The VLM is given a natural image of a real-world system (e.g., cities, clouds, vegetation) and is tasked with describing the system and writing code that simulates and generates it. This generative code is then executed to produce a synthetic image, which is compared against the original. This approach is tested on various complex emergent systems, ranging from physical systems (waves, lights, clouds) to vegetation, cities, materials, and geological formations. Through analysis of the models and images generated by the VLMs, we examine their understanding of the systems in images. The results show that leading VLMs (GPT, Gemini) have the ability to understand and model complex, multi-component systems across multiple layers of abstraction and a wide range of domains. At the same time, the VLMs exhibit limited ability to replicate fine details and low-level arrangements of patterns in the image. These findings reveal an interesting asymmetry: VLMs combine high-level, deep visual understanding of images with limited perception of fine details.

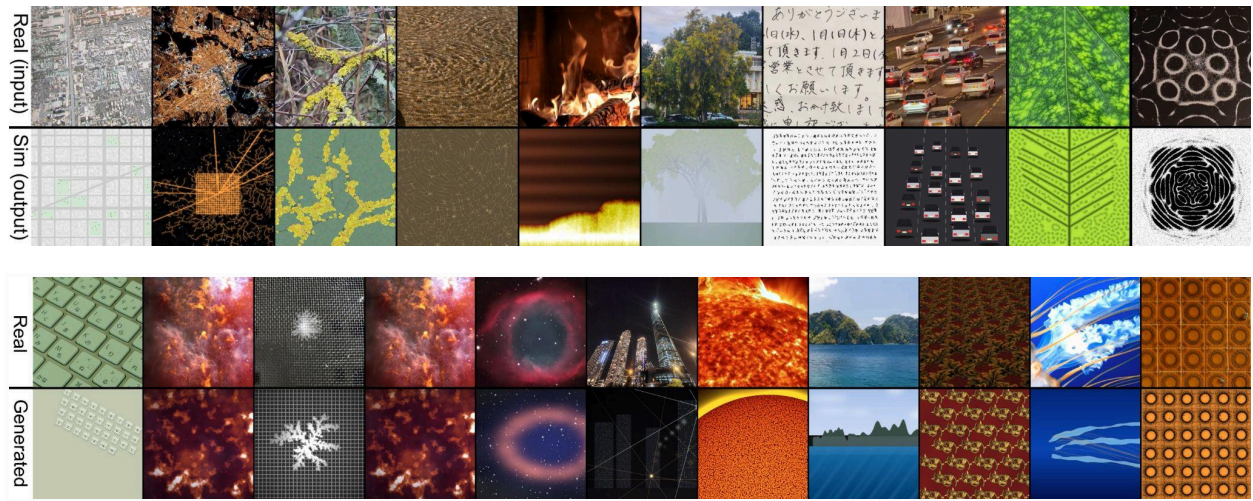

**Figure 1)** The Im2Sim approach. The Vision Language Model (VLM) is given an image of a real-world system (Real) and is asked to describe the process that forms the pattern in the image and write code that simulates the process. The code is executed to generate a synthetic image (Sim/Gen). Each generated/simulated image corresponds to the real image directly above it.

---

[1] Weizmann Institute AI Hub, sagieppel@gmail.com
\* Code and data are available [at this url](#)

# 1. Introduction

Understanding any phenomenon is closely related to the ability to create a representative model of that phenomenon. Similarly, the ability to construct a representative model of a system shown in an image represents one of the deepest forms of visual understanding[1-8]. While tasks like image classification, segmentation, and pattern recognition can be accomplished using simple algorithms[9,10], creating a model or simulation of a system in an image requires genuine understanding of the underlying mechanisms and processes that formed this system. Vision Language Models (VLMs) have emerged as the leading approach for general visual intelligence. These models are trained on vast amounts of visual question-answering (VQA) and image-to-text data, enabling them to analyze and answer complex questions about images across a wide range of fields[10-13], yet at the same time fail at seemingly trivial tasks[14-16]. This has led to ongoing debate about whether VLMs truly possess world understanding. This work explores VLMs' capacity to understand and simulate the generative processes underlying real-world visual patterns using the **Im2Sim2Im** method (Figure 2): The VLM receives a natural image depicting an emergent pattern, which can be physical, biological, chemical, or even social systems like clouds or waves, vegetation, colonies, cities, materials, or geological formations. The model is tasked with identifying the mechanism that created the pattern and writing code to simulate this process. The simulation code is then executed to generate a synthetic image that can be compared with the original input image. The Im2Sim[17] method produces three outputs (process description, simulation code, and synthetic image), which are analyzed to evaluate VLMs' understanding of the underlying system. The results demonstrate that VLMs can understand physical, biological, and social systems across multiple layers of abstraction, creating valid simulations of the systems that form these images. Beyond evaluation, the Im2Sim approach generates a diverse collection of pattern and texture generators useful for any application requiring such assets[16].

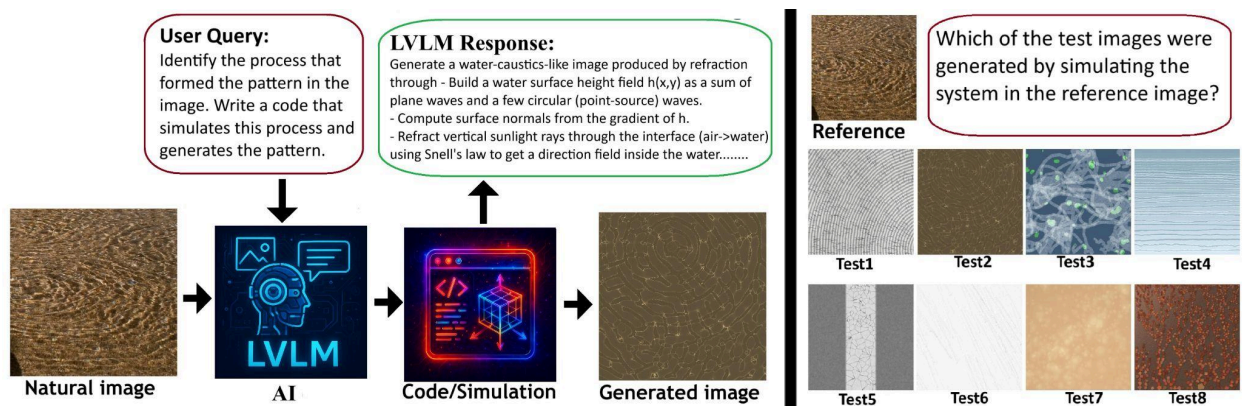

Figure 2: The Im2Sim2Im approach for testing VLM's ability to identify and model the underlying mechanism behind a real-world visual pattern. Left: The VLM receives an image of a real-world pattern, infers the physical process that formed this pattern, implements it as code, and runs the code to generate a simulated image. Right: A matcher (either human or VLM) evaluates and ranks the simulated image by identifying which test image best matches the reference image containing the real-world pattern. The hypothesis is that the more accurate the simulation is, the more similar the resulting image will be to the input image.

.

## 1.1. Im2Sim Quantitative vs Qualitative evaluation

The Im2Sim method (Figure 2, left) allows both quantitative and qualitative evaluation of VLM abilities. Quantitative evaluation was discussed in previous work[17] and is done by matching the image generated by the simulation to the original image. This is based on the assumption that the more accurate the simulation, the more similar the generated image will be to the input real-world image (Figure 1,2). This similarity can be evaluated using a multiple-choice image matching question (Figure 2, right): Human or VLM evaluators are presented with the original image as reference, while the synthetic image generated by simulating the original image, and 9 decoy images generated by different simulations are used as choices. The evaluator is tasked with identifying which generated image best matches the system of the reference image (Figure 2, left). If the simulation is accurate, the generated image will be similar to the input image, and the two will be easily matched (Evaluator is accurate and will match the most similar images). The results of this evaluation method are given in Table 1. These results support the idea that VLMs can indeed create valid models of systems, with all VLMs showing an accuracy of 50%-80% compared to 10% at random. This method is discussed in more detail in [17]. In contrast, this work focuses on qualitative evaluation by examining different models and simulations generated by VLMs in response to images of various systems, directly comparing the original image with the generated code and simulated images. The work is divided into sections discussing various systems: physics-based, plants and vegetation, cities and settlements, text and symbols, patterns and tiling.

**Table 1:Im2Sim2Im results (Figure 2): The VLM is given a natural image and is asked to identify the process that generates the pattern in the image, simulate it using code, and run this code to generate a simulated image (Figure 2.lef). The matching accuracy of the simulated image to the input natural image is used to evaluate the model accuracy (Figure 2 right). Matching was done with a single real image as a reference and 10 synthetic test images, each made by a different model. The image matching/ranking (Figure 2.left) of Im2Sim2Im was done using GPT-5 with color or grayscale images, or by a human evaluator (with color images). For more details, see previous work[17].**

| Evaluator | GPT-5 | GPT-5 mini | Gemini 2.5-flash | Gemini 2.5-pro | Qwen2.5 VL-72B | Llama-4 Maverick17B | Llama-4-Scout | Grok-4-fast | Grok-4 reasoning |
|---|---|---|---|---|---|---|---|---|---|
| Color | 74 | 77 | 79 | 78 | 41 | 54 | 54 | 62 | 71 |
| Gray | 62 | 70 | 64 | 67 | 35 | 47 | 46 | 62 | 61 |
| Human | 82 | 75 | 74 | 68 | 45 | 60 | 61 | 70 | 78 |

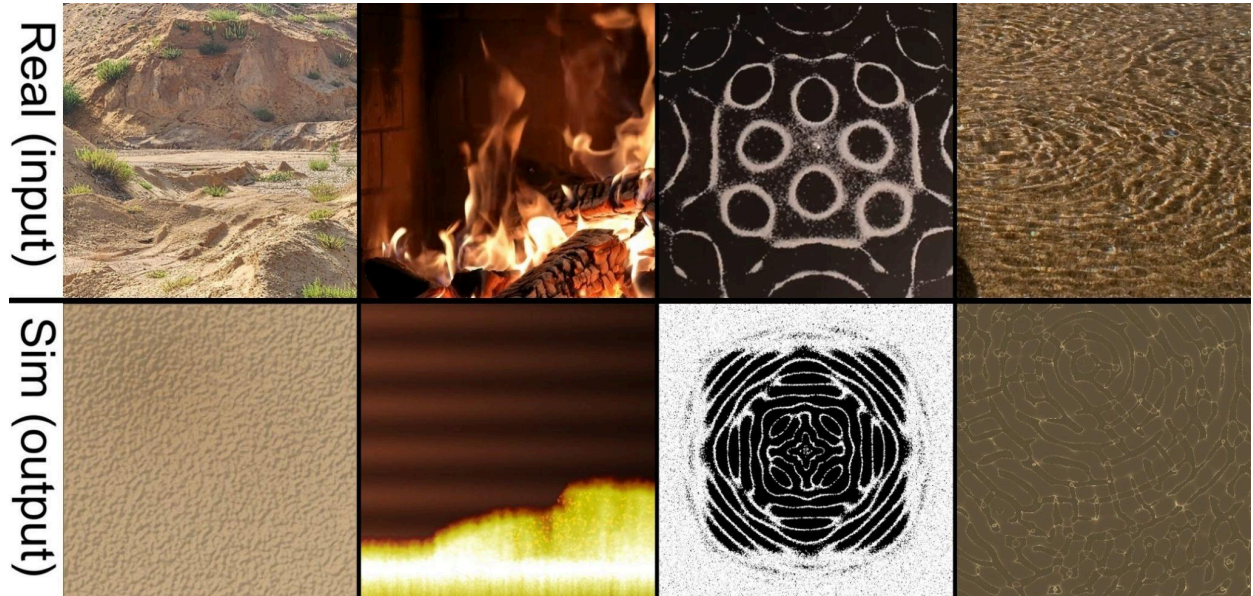

**Figure 3: Physical phenomena. Real (Top):** Real images given as input to the VLM. **Sim (Bottum):** Simulated image created by VLM modelling of the system in the real image (the image directly above it).

## 2. Physical patterns

When dealing with images of physical phenomena, most VLMs were able to recognize the underlying processes and represent them across multiple physical domains. Particularly impressive is their ability to identify distinct components of a system and model each using different approaches. Modeling complex systems often requires trade-offs: some components are represented with coarse approximations, while others are simulated with higher fidelity. For example, waves on water can be modeled through detailed fluid dynamics or approximated as simple sinusoidal waves. Similarly, a crowd of people may be represented as interacting intelligent agents or as a statistical random walk. Deciding which part of the system to model using detailed simulations versus simple approximation is a central challenge. For example, when simulating the distribution of particles on a vibrating plate under standing sound waves (Chladni plates, Fig 3. Panel 3), the VLM (GPT-5) represents the standing wave as a superposition of sinusoidal functions, thereby bypassing the complex particle dynamics simulation. Instead, it assumes a Boltzmann-like distribution of particle positions over the wave field, which is then sampled to scatter particles and generate the image. While these approximations do not fully reflect the underlying physics, they successfully capture the high-level spatial distribution of particles. However, the VLM clearly fails to identify the specific combinations of wave modes responsible for a given Chladni pattern, producing incorrect standing-wave spatial patterns. Similarly, when given an image of light reflected from a wavy water surface (Fig 3. Panel 4), the VLM simulates the wavy water surface as a simple sinusoidal function, but then applies accurate physics using Snell's law to calculate the reflectance of light from the water surface, capturing both the wave and the caustic reflection patterns. Again, this highlights the VLM's ability to model the system at multiple levels of abstraction, ranging from rough approximations (a sinusoidal wave for the water surface) to accurate physics (Snell's law for light). For geological structures of weather-eroded dune (Fig 3. Panel 1), the VLM creates a digital sandbox that simulates surface formation, rain, water flow, erosion, and sedimentation using a multicomponent cellular model that combines all of these aspects into a single simulation. While this is a simplified toy model that captures only the general principles and the generated image was far from the input image, it captured the core physics of the system and the pattern of the image. When generating a fireplace flame image (Fig 3. Panel 2), the VLM represents the fire as a simplified 2D heat field and models combustion through stochastic heat injection near the bottom of the domain, with enhanced intensity around static log-shaped obstacles. Rather than explicitly solving fluid dynamics, it approximates buoyant flow by advecting heat upward with a velocity proportional to local temperature. The heat field was then diffused and subjected to height-dependent cooling to approximate energy loss as the flame rises. Again illustrates how the VLM identifies system components and simulates them, combining coarse physical approximations (a scalar heat field and heuristic buoyancy) with more principled numerical methods (semi-Lagrangian advection and diffusion). In approximately half of the cases, the VLM ignores the prompt to explicitly simulate the system physics and instead attempts to directly replicate patterns observed in the image. Interestingly, in many instances, this strategy produces images that are more similar to the real input, as it gives the model greater control over the physical appearance of the output. This form of "cheating" is a major limitation of the Im2Sim2Im approach, and is the reason why some weaker (mini) models rank better on the qualitative evaluation (Table 1).

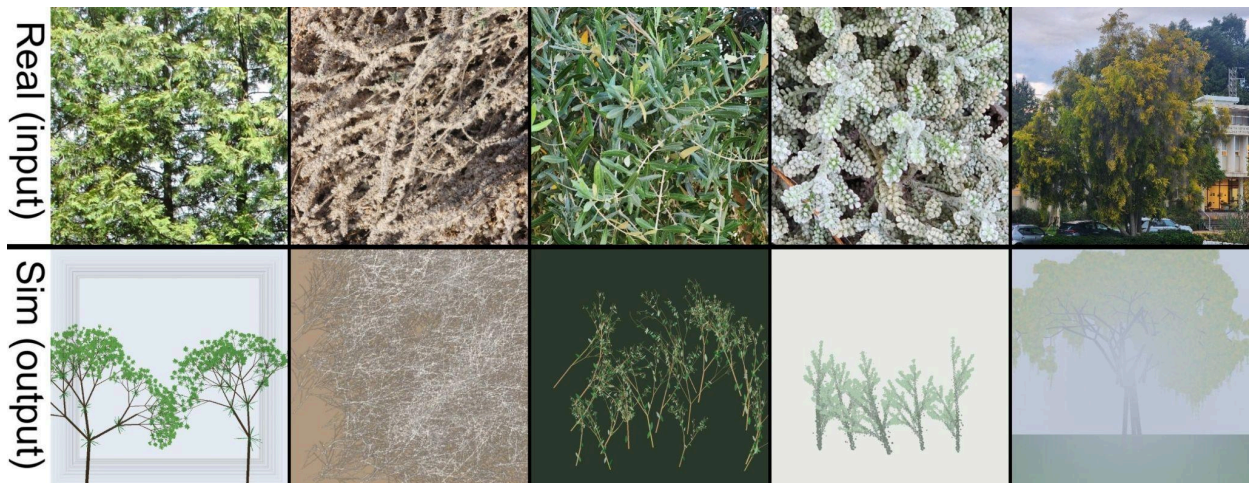

**Figure 4: Branching vegetation and L-System model. Real (Top):** Real-world images given as input to the VLM. **Sim (Bottum):** Generated image, created by VLM modelling of the system in the real image.

## 3. Trees and vegetations:

For images of vegetation and trees, models such as GPT and Gemini rely on a combination of methods, including L-systems[18,19] for branching structures and reaction-diffusion or fractal

Brownian motion (fBM)[20] for leaves and bark patterns. These are well-known modeling approaches, but they are highly general and strongly dependent on parameters and branching rules. L-systems, in particular, are extremely flexible and sensitive to branching definitions, requiring the VLM to identify branch and leaf structures and encode them into appropriate branching rules. As can be seen in the generated images (Figure 4), while the resulting structures are clearly rough, they capture the overall branching patterns and general leaf shapes. Interestingly, despite the VLM's ability to reason about the structure at multiple levels, it typically focuses on modeling a single dominant aspect (e.g., branches, leaves, or overall structure), while representing other components using much simpler approximations. For example, the model may use a tuned L-system to generate branching, but represent leaves as simple ellipses (Figure 4). Leaf and bark textures are sometimes modeled using reaction-diffusion processes or fractal Brownian motion; however, in these cases, the VLM often fails to capture the overall visual appearance (Figure 5). This is likely because the relationship between the input parameters of these models and the resulting visual patterns is significantly less predictable.

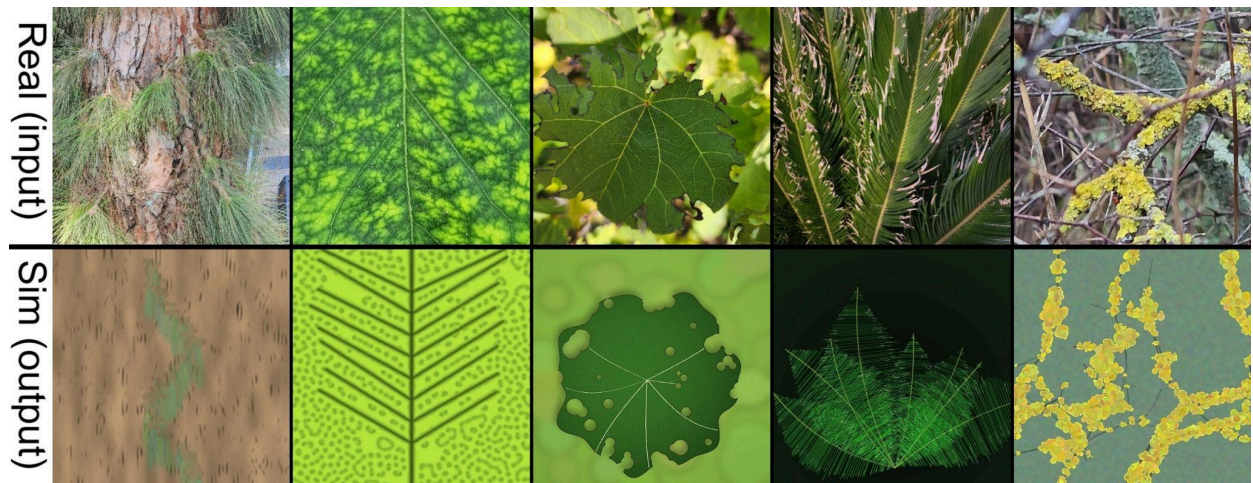

**Figure 5: Vegetation models. Real (Top): Real-world images given as input to the VLM. Sim (Bottum): Generated image, created by VLM simulating the system in the real image.**

# 4. Cities and settlements:

Images of cities and settlements present a unique challenge due to the complex geographic, economic, and social systems underlying their formation. VLMs managed to identify core urban characteristics and growth patterns, such as grid versus organic street networks, and key features like monocentric versus polycentric development or sparse versus dense layouts, and selected simulation models accordingly (Figure 6). The modeling approach often employed multi-layered simulation, where each layer guided the formation of subsequent ones. For example: First, geography (land/sea boundaries) was created using procedural functions like Perlin noise. This geographic layer then guided population distribution and urban centers through network growth models[21,22], while more ordered cities utilized grid subdivision algorithms (Fig 6.right). Population density, in turn, directed road network formation via preferential attachment or space colonization algorithms. Each layer (geography, population, road networks, lighting) required distinct yet interdependent models and parameters guided by its predecessor. While VLMs generally selected appropriate models for each subsystem (predominantly network growth approaches), the parameter tuning and model integration proved imprecise at best (Fig.6). Results captured much of the high-level general structure but exhibited numerous anomalies and almost no spatial correspondence to input city (Fig.6). This suggests that VLMs recognize and understand cities at a broad, conceptual level, but have difficulty recognizing and representing the finer details, low-level patterns, and spatial arrangements, indicating a gap between conceptual comprehension and detailed low-level pattern recognition.

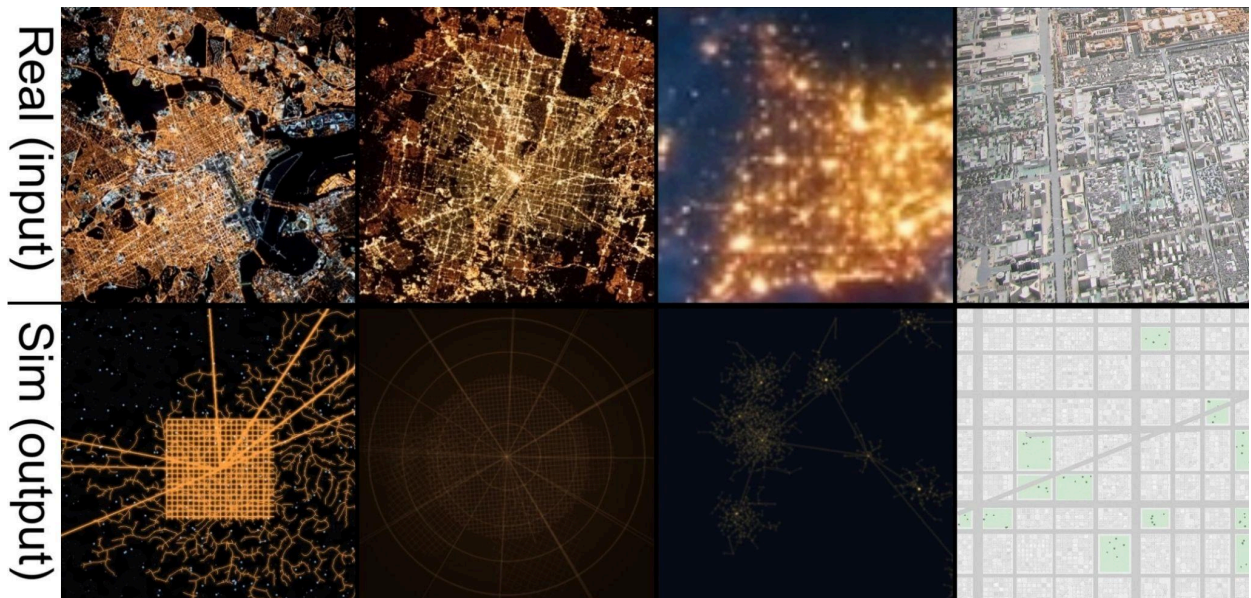

**Figure 6: Cities and settlements. Real (Top): Real images given as input to the VLM. Sim (Bottum): Simulated image, created by VLM modelling of the system in the real image.**

# 5. Text and symbols

The way VLMs approach modeling text and scripts is particularly revealing due to the multiple layers of abstraction involved. When shown an image of handwriting (Fig. 7, panel 3), the VLM (GPT-5) chose to simulate individual brush strokes that form the characters, creating a page of convincing-looking handwritten symbols that matched the general calligraphic style but carried no actual meaning. In contrast, when given a printed page, the VLM essentially replicated the printing process word by word using standard fonts while also attempting to replicate page textures (Fig. 7, panel 1). Other models took different approaches: some generated random letter-like shapes with basic visual similarity using Markov chains. Amusingly, when given a doodle board, GPT5 responded with its own set of doodles (Fig. 7, panel 4). In all cases, the VLMs seem to understand the underlying object (book, board, keyboard), the letter structures (strokes, fonts, doodles), and the context, although often not all at the same time. Two notable outliers emerged: Grok created an image of a written page but replaced the text with a prompt instructing itself to write the model. Llama, meanwhile, simply generated a page of blank lines, perhaps the most minimalist interpretation of a text page possible.

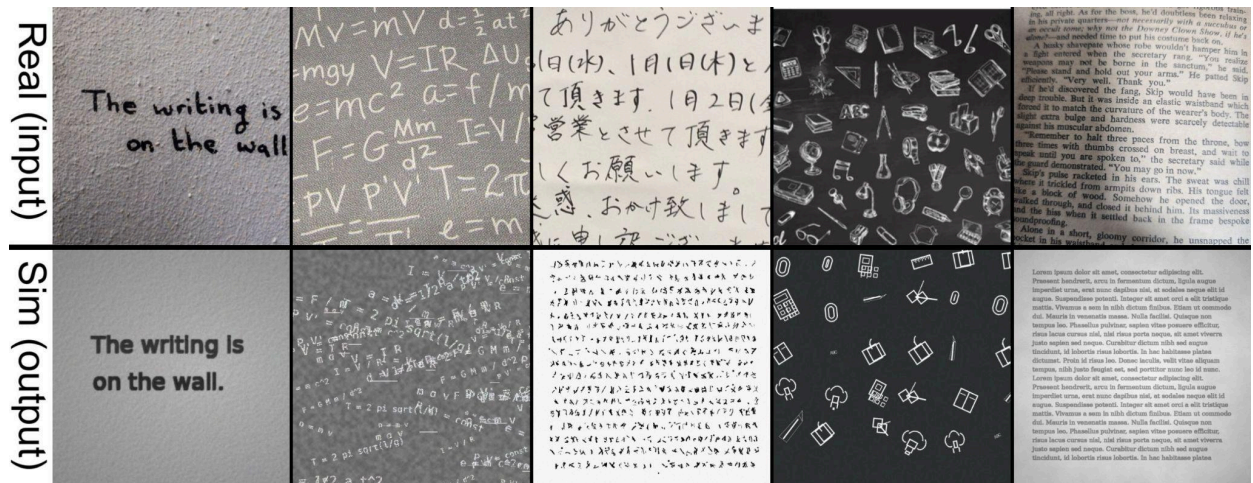

**Figure 7: Text and script. Real (Top): Real images given as input to the VLM. Sim (Bottum): Simulated image, created by VLM modelling of the system in the real image.**

# 6. Modelling visual patterns

In many cases, VLMs recognized the underlying physical system but deliberately chose to bypass the actual physics mechanisms, instead creating simplified functions to replicate the visual outcome. This pattern emerged consistently when dealing with complex systems that are computationally expensive or difficult to simulate, and was more common in small (mini/flash) VLMs. In these instances, VLMs behave more like artists combining multiple visual functions to achieve the desired appearance rather than scientists attempting to capture the underlying physics. Consider the sun's surface granular cells, which form through fluid and plasma convection processes (Fig. 8, panel 6). While simulating the actual convection would be computationally prohibitive, GPT-5 approximated these patterns using modified Voronoi cells, a far simpler approach that captures the visual essence. The same strategy appeared in cloud and galaxy formation, where VLMs entirely sidestepped physical simulation in favor of general procedural functions like fractional Brownian motion (fBM) and Perlin noise that together replicate the general pattern. While such approaches ignore the physical mechanisms, they reveal an impressive ability to recognize and reproduce visual structures using code (Fig. 8).

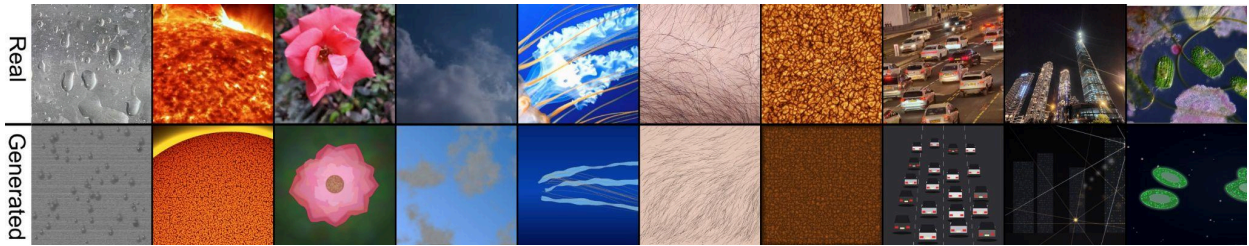

**Figure 8: Various cases where the VLM tried to replicate the visual pattern without simulating the physical system. Real (Top): Real images given as input to the VLM. Sim (Bottum): Image, created by VLM trying to replicate the pattern in the real image.**

For static patterns like floor tiling, repetitive textiles, and stairs, most VLMs easily identified and replicated the underlying structure with good accuracy for simple patterns, even for 3D structures (Fig. 9). The VLMs show the ability to model the system in multiple layers, capturing not only the core pattern but also cracks and imperfections (Fig. 10). However, when the core tiling pattern was more complex , such as in cases of complex tapestry, all VLMs often failed to capture the exact underlying structure, instead generating patterns that contained similar sets of shapes and colors but lacked the fine-grained details (Fig. 9).

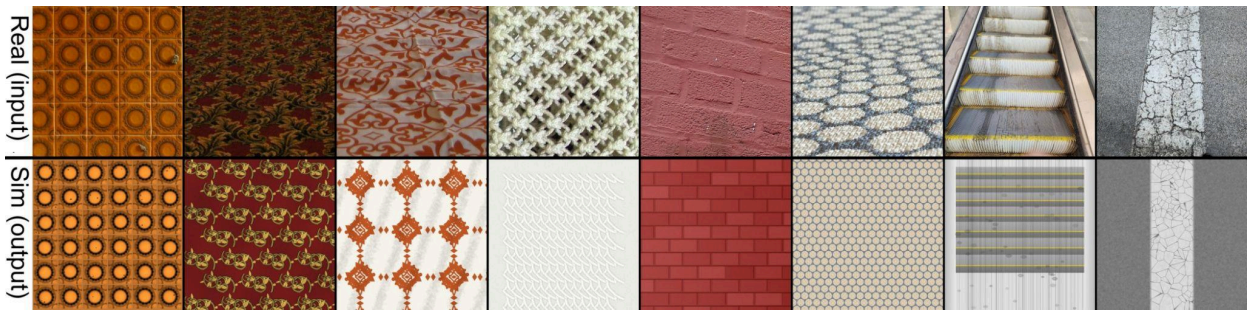

**Figure 9: Tiling and repetitive patterns. Real (Top): Real images given as input to the VLM. Sim (Bottom): Image, created by VLM trying to replicate the pattern in the real image.**

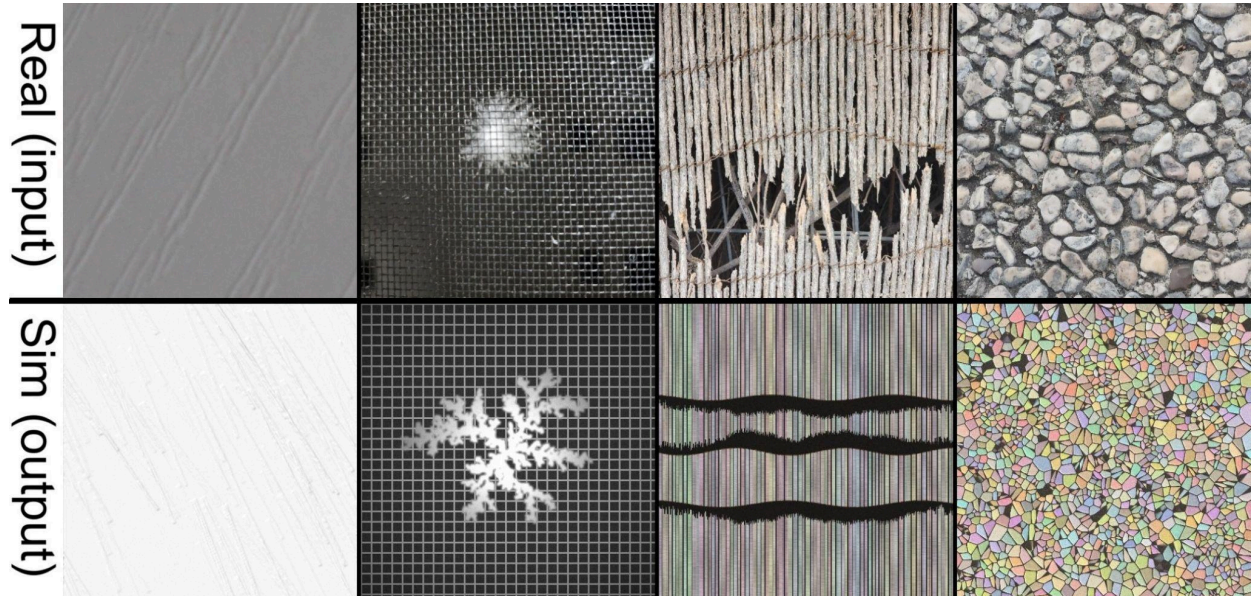

Figure 10: Tiling and repetitive patterns with imperfections. Real (Top): Real images given as input to the VLM. Sim (Bottum): Image, created by VLM trying to replicate the pattern in the real image.

# 7. Conclusion

The main picture emerging from this work is that vision-language models (VLMs) are capable of understanding and modeling systems across a wide range of domains. VLMs, such as GPT-5 and Gemini-2.5, were not only able to identify the core patterns and mechanisms present in images but also to approximate and simulate them in ways that capture the dominant structures and dynamics. These simulations reproduce the underlying physical systems and processes that generate the observed patterns, providing strong evidence that VLMs exhibit a form of deep understanding of the images they process. Examination of the generative code and models produced by VLMs reveals a clear grasp of the key aspects of the underlying systems. Particularly impressive is the ability to decompose a system into distinct components and to simulate each at different levels of abstraction, either by modeling the underlying physical mechanism directly or by approximating it with a simpler function that reproduces the core patterns. Although the resulting simulations are often rough, they nonetheless demonstrate that VLMs can achieve a form of deep mechanistic understanding across a broad range of systems. The models and simulation strategies employed by VLMs are typically based on well-known frameworks (e.g., Brownian motion, Boltzmann distributions, L-systems). However, the ability to select appropriate models, set their parameters and rules, and combine multiple models within a single simulation still requires substantial understanding of the system as well as the ability to identify and understand main patterns in the image (leaf shapes, branch structure), and the way different processes interact and affect each other. At the same time, VLMs show clear limitations in capturing low-level aspects of images. They frequently miss fine-grained spatial arrangements, fail to reproduce exact visual patterns, and struggle to correctly integrate multiple components to form specific patterns. This suggests an intriguing asymmetry: VLMs demonstrate strong high-level understanding of the generative structure of systems while remaining limited in their ability to perceive and represent fine visual details of those same systems.